\documentclass[conference,a4paper]{IEEEtran}
\IEEEoverridecommandlockouts
\usepackage{cite}
\usepackage{amsmath,amssymb,amsfonts}
\usepackage{algorithmic}
\usepackage{graphicx}
\usepackage{textcomp}
\usepackage{xcolor}
\def\BibTeX{{\rm B\kern-.05em{\sc i\kern-.025em b}\kern-.08em
    T\kern-.1667em\lower.7ex\hbox{E}\kern-.125emX}}
\begin{document}

\title{Improving Industrial Injection Molding Processes with Explainable AI for Quality Classification
\\
}

\author{\IEEEauthorblockN{1\textsuperscript{st} Georg Rottenwalter}
\IEEEauthorblockA{
\textit{Rosenheim Technical University of Applied Sciences} \\
Rosenheim, Germany \\
georg.rottenwalter@th-rosenheim.de}
\and
\IEEEauthorblockN{2\textsuperscript{nd} Marcel Tilly}
\IEEEauthorblockA{\textit{Rosenheim Technical University of Applied Sciences} \\
Rosenheim, Germany \\
marcel.tilly@th-rosenheim.de}
\and
\IEEEauthorblockN{3\textsuperscript{rd} Victor Owolabi}
\IEEEauthorblockA{\textit{Rosenheim Technical University of Applied Sciences} \\
Rosenheim, Germany \\
victor.owolabi@stud.th-rosenheim.de}
}

\maketitle

\begin{abstract}
Machine learning is an essential tool for optimizing industrial quality control processes. However, the complexity of  machine learning models often limits their practical applicability due to a lack of interpretability. Additionally, many industrial machines lack comprehensive sensor technology, making data acquisition incomplete and challenging. Explainable Artificial Intelligence offers a solution by providing insights into model decision-making and identifying the most relevant features for classification. In this paper, we investigate the impact of feature reduction using XAI techniques on the quality classification of injection-molded parts. We apply SHAP, Grad-CAM, and LIME to analyze feature importance in a Long Short-Term Memory model trained on real production data. By reducing the original 19 input features to 9 and 6, we evaluate the trade-off between model accuracy, inference speed, and interpretability. Our results show that reducing features can improve generalization while maintaining high classification performance, with an small increase in inference speed. This approach enhances the feasibility of AI-driven quality control, particularly for industrial settings with limited sensor capabilities, and paves the way for more efficient and interpretable machine learning applications in manufacturing.
\end{abstract}

\begin{IEEEkeywords}
industry, injection molding, explainable ai, machine learning, time series data
\end{IEEEkeywords}

\section{Introduction}
The use of machine learning (ML) in industrial processes has increased significantly in recent years, particularly in quality control. ML models are already being used to identify defective products and reduce rejects by using data-driven quality models for process optimization \cite{rai2021machine}. Although numerous powerful methods exist, one central problem remains: the lack of interpretability of the models. This not only makes it difficult to trust the predictions but also to analyze errors and improve processes.

 To close this interpretability gap, Explainable Artificial Intelligence (XAI) offers promising approaches to make the decision-making of ML models more transparent. XAI techniques make it possible to understand the underlying mechanisms of quality models better and to analyze the relevant features of a process in a more targeted manner. This not only makes decisions more comprehensible but also allows the causes of quality defects to be identified more precisely \cite{gadekallu2024xai}. This will help to make industrial processes more efficient and increase confidence in AI-supported quality controls.

 In this context, various XAI methods were used in this thesis to understand better the factors influencing quality prediction in the field of injection molding. 
Specifically, the three popular XAI methods SHAP, Grad-CAM, and LIME were used to analyze the significance of the individual input parameters from our time series data of an Long Short-Term Memory (LSTM) model for quality classification \cite{nguyen2021evaluation}. 
For the use in our data scenarios, we further refer to the work by Theissler et al. in `Explainable AI for Time Series Classification: A Review, Taxonomy and Research Directions', \cite{theissler2022explainable} which highlights these methods as established post-hoc explainability techniques suitable for time series classification. While Grad-CAM was originally developed for Convolutional Neural Networks (CNNs), we adapted it to our LSTM model by defining the hidden state outputs as the representation to be explained.
 The aim was to reduce the original 19 training parameters by half and by two-thirds to nine and six relevant features. This selection was used to analyze whether a more efficient and simpler process classification is possible, without accepting significant losses in model performance.

The main objective of this work was to reduce the input parameters for quality classification in injection molding using XAI techniques. The aim was not only to reduce the number of features but also to identify those parameters with the highest relevance for the prediction quality. In both research and industry, many production facilities rely on older or less well-equipped machines that lack comprehensive sensor technology for complete data acquisition. This poses a challenge for AI-based quality control, as missing sensor data can limit model performance. The approach presented here aims to identify the most critical parameters for quality classification and to evaluate whether a targeted reduction to six features can maintain comparable model performance. In the long run, this could not only optimize data acquisition, but also make AI-based quality control more feasible for existing industrial systems by focusing on the most relevant sensor values.

A central aspect was to analyse whether models can be trained with the reduced parameters that achieve a similar accuracy or F1 score in classification as models with the full feature set. At the same time, it was tested whether the inference speed of the models is improved by reducing the input variables, which is of particular importance for real-time applications in industrial quality control.

Another important point was the use of synthetic data for model development. As it is difficult to generate realistic simulator data with a complete set of 19 input features \cite{georg2023advancements}, it was investigated whether this process could be simplified through targeted feature reduction. For example six essential parameters could make the generation of synthetic training data more efficient and improve the applicability of simulation data for the training of ML models.

\section{Related Work}

\subsection{Explainable AI background}

Understanding how machine learning models make decisions is crucial, especially in industrial applications where transparency is required. The following approaches provide an overview of widely used XAI methods that help interpret and analyze model predictions.

An important approach to the interpretability of machine learning models is presented in `A Unified Approach to Interpreting Model Predictions' \cite{lundberg2017unified}, in which the SHAP (SHapley Additive exPlanations) method is developed. The authors combine concepts of game theory with XAI techniques to quantitatively determine the influence of individual input features on model predictions. SHAP offers a standardized and consistent method for attributing model decisions and is frequently used for feature analysis in complex industrial applications.

Another contribution to the field of XAI is Grad-CAM, a method developed by Ramprasaath et al. \cite{ramprasaath2019visual}. Their approach enables visualization of the neural network decision-making process by identifying activated image regions that contribute substantially to classification. In the particular context of convolutional neural networks (CNNs), Grad-CAM is used to enhance the comprehension of the functionality of the model and to ensure greater transparency in their predictions.

Another approach to the explainability of classification models is presented by Ribeiro et al. \cite{ribeiro2016should}. The authors develop LIME (Local Interpretable Model-agnostic Explanations), a method that makes black-box models more comprehensible through local approximation with interpreted, linear models. Thanks to its flexible applicability to different classifiers, LIME helps to make individual predictions comprehensible and thus enables transparent analysis of model decisions.

Theissler et al. provide a comprehensive perspective on explainable artificial intelligence (XAI) in the context of time series in their paper, `Explainable Artificial Intelligence for Time Series Classification: A Survey' \cite{theissler2022explainable}. In their survey, the authors offer a structured taxonomy of XAI methods for time series classification, categorising them as time-point, subsequence, or instance-based approaches. They highlight the increasing importance of interpretability in time-dependent data domains and identify relevant research gaps, such as the absence of user studies and appropriate evaluation metrics.

\subsection{Explainable AI in the industry}

While general-purpose XAI methods provide valuable insights into model behavior, their applicability in industrial settings presents additional challenges. The following studies illustrate how XAI has been integrated into industrial processes to improve transparency, efficiency, and confidence in AI-driven decisions.

The use of XAI in industrial applications is investigated by Lin et al \cite{lin2024explainable}. The authors combine XAI with a multi-stage transfer learning (TL) approach and artificial neural networks (ANN) to accurately predict the quality of injection-molded polycarbonate lenses. By integrating simulations and real production data, the model enables more efficient process development, reduces the need for extensive training data, and improves the traceability of predictions. This work shows how XAI can help to make industrial manufacturing processes more transparent while increasing the efficiency of quality prediction.

The use of XAI in the process industry is analyzed in Kotriwala et al. \cite{kotriwala2021xai}. The authors discuss the challenges and requirements of applying XAI in highly automated industrial processes, where human decision-making continues to play a central role despite AI-supported optimization. The authors argue that successful integration of XAI cannot be based on algorithmic improvements alone, but must also take into account domain-specific knowledge and human factors. The paper identifies research gaps and shows that XAI can be a crucial tool for transparency and acceptance of AI in industry.

A comprehensive overview of XAI is given in `Peking Inside the Black-Box: A Survey on Explainable Artificial Intelligence (XAI)' \cite{adadi2018peeking}. The authors address the growing importance of XAI in view of the increasing spread of AI systems and their often lack of transparency. A detailed examination of the existing explanation methodologies is conducted, with a focus on identifying prominent research trends and articulating the potential of XAI to enhance the reliability of AI-driven systems. This study functions as an introductory guide for researchers and practitioners, underscoring the necessity for ongoing advancements in this rapidly evolving domain of research.

\subsection{AI in injection molding}

The integration of AI into injection molding is an active area of research aimed at improving process stability and product quality. The following approaches illustrate how machine learning and AI techniques have been specifically used to optimize injection molding processes and improve predictive quality assessment.

An overview of the use of machine learning in injection molding is given in `A Review on Machine Learning Models in Injection Moulding Machines' \cite{selvaraj2022review}. The authors analyze various ML approaches, in particular neural networks, for optimizing process parameters such as injection speed, pressure, and mold design. The paper also discusses challenges such as data preparation, model validation and the practical application of ML methods for quality improvement in industry.

An AI-supported approach to optimizing the injection molding process is presented by Park et al \cite{park2019ai}. The authors develop a real-time system that uses pressure and temperature sensors to monitor process parameters and use machine learning algorithms to automatically make adjustments in order to minimize quality deviations. By combining mold flow simulations and real industrial data, the work shows that adaptive control can improve product consistency.

Another relevant approach is presented in `Cross-Machine Predictions of the Quality of Injection-Moulded Parts' by Combining Machine Learning, Quality Indices, and a Transfer Model \cite{chang2024cross}. The authors develop a model that can predict product quality across different injection molding machines using quality indices and a transfer model. By combining real machine data with simulations, the work shows that precise quality prediction across different machines is possible and production deviations can be efficiently compensated.

\section{Methods}

Figure 1 shows the system structure of our approach for identifying the most important parameters in the injection molding process using XAI. The concept comprises two central steps: the evaluation of the model with real production data and the analysis of the most important features to optimize the training.

\subsection{Training Evaluation Data}
To identify the most important parameters for quality prediction in the injection molding process, a three-layer LSTM network was developed for time series data from machine processes. The model was trained with real, labeled production data collected during running production cycles. The input data included various sensor values such as temperature, pressure, and injection speed over the entire production cycle. By analyzing the model performance on a separate test data set, it was verified that the network was able to perform a reliable quality classification.

\subsection{Parameter analysis}
After the initial model training, XAI methods such as SHAP, Grad-CAM, and LIME were used to determine the most influential input variables of the LSTM model. Through this analysis, the parameters that had the greatest influence on the quality classification were identified. The model was then retrained with the reduced set of parameters to investigate whether a comparable classification accuracy could be achieved with a smaller number of features. In addition to evaluating the model performance in terms of accuracy and F1 score, it was also analyzed whether the inference speed improved as a result of the reduced input variable.

\begin{figure}
\includegraphics[width=\columnwidth]{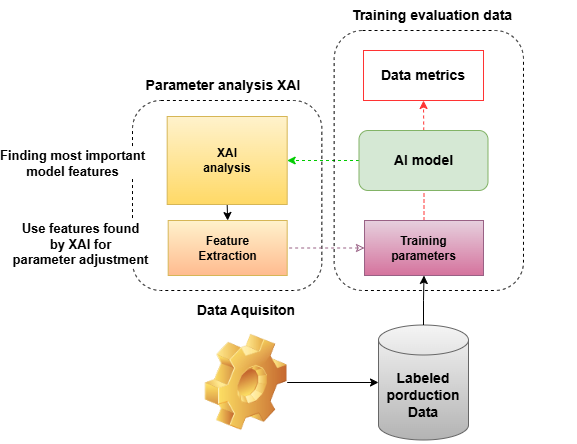}
\caption{System design of overall concept.} \label{fig1}
\end{figure}

\section{Experimental setup}
\subsection{Training \& validation data}
A data set of 1,171 evaluated injection molding cycles using polypropylene as the material was used for our study, where a small plastic box measuring approximately 10 × 10 × 15 cm was molded as the cavity and mold. An experienced injection molding operator evaluated the quality of the parts produced and documented relevant process data for each production cycle. To ensure broad process variability, key process parameters were specifically varied, including piston stroke, rotations per minute during dosing, injection volume, holding pressure, holding time, and other relevant factors.

Each cycle comprises around 1800 time steps, with the sensor data recorded at a fixed 10 ms interval. To improve the training of the model, a data augmentation was applied to increase the diversity of the time series data. We enlarged each dataset by quadrupling it and stored each first through the fourth step at 10 ms intervals in separate files.

The data set was split with a train test split into a ratio of 33\% validation data and 67\% training data. This resulted in 3138 training datasets and 1546 validation datasets that were randomly intermixed.

\subsection{LSTM Training}

The LSTM model for our neural network was specially developed for processing time series data in the injection molding process. It consists of three LSTM layers, each with downstream dropout layers to avoid overfitting. The first two LSTM layers are set to True with the \textit{return\_sequences} attribute so that they can retain their state over several time steps. The architecture was based on the concept from `Sequence to Sequence Learning with Neural Networks' \cite{sutskever2014sequence}, which demonstrates the effectiveness of deep LSTM networks in handling complex tasks. Our LSTM network follows a unidirectional many-to-one configuration, whereby a sequence of input data is processed and a single classification output is generated \cite{smagulova2019survey}.

Three different feature sets were used for the training runs:

\begin{itemize}
\item 19 features (complete data set)
\item 9 features (halved XAI data set)
\item 6 features (XAI data set reduced by 2 thirds)
\end{itemize}

These variations made it possible to analyze the influence of the input features on the model performance and to evaluate whether a reduction of features can improve classification accuracy and inference speed.

Table \ref{tab:train_parameters} is providing an overview of the configuration of the LSTM network, including the layer structure, activation functions, and other hyperparameters. The choice of these parameters was optimized through hyperparameter tuning experiments to achieve the best possible model performance.

All calculations were performed on a system with a 12th Gen Intel Core i9-12900KS 3.40 GHz CPU, 64GB of RAM, and an NVIDIA GeForce RTX 3090 Ti GPU. TensorFlow 2.5.3 were used as the development environment.

\begin{table}
\centering
\caption{Variables for network training that have remained constant}\label{tab1}
\begin{tabular}{|l|l|l|}
\hline
Learning rate & 0.001325\\
Input dimension & 19\\
Output size & 1\\
Units LSTM layer 1 & 300\\
Units LSTM layer 2 & 100\\
Units LSTM layer 3 & 100\\
Dropout rate behind LSTM layer 1 and 2 & 0.20181\\
Dropout rate after last LSTM layer & 0.17249\\
Optimizer & Adam\\
Loss & binary cross-entropy\\
Batch size & 64\\
Epochs & 350\\
\hline
\end{tabular}
\label{tab:train_parameters}
\end{table}

\subsection{XAI parameter analysis}

We applied SHAP, Grad-CAM, and LIME to identify the best input features in our LSTM model.

For SHAP (version 0.42.1), we use the GradientExplainer, as it is compatible with TensorFlow 2.5.3 and offers CUDA core support. This enables efficient calculation of feature weights on the GPU.

Grad-CAM is used to identify the most active areas of LSTM cells during the prediction process. As there is no official package for sequential models, we have developed a custom implementation to analyze gradient-based activation patterns. Taking inspiration from the original Grad-CAM for CNNs \cite{selvaraju2017grad} and its extension to 1D-CNNs for text classification \cite{gorski2020towards}, we apply gradient-weighted relevance to the hidden states of recurrent architectures. We compute the gradient of the predicted class score with respect to the LSTM hidden states, average these gradients over time, and use them to weight the activations. Collapsing the resulting heatmap across time yields a feature-level importance representation that is class-discriminative and reveals the model’s internal focus. This method requires no architectural changes and can be applied to any differentiable sequential model with recurrent components.

For LIME (version 0.2.0.1) we use the LIME-Tabular Explainer. To determine the global importance of the features, we average the values of each input variable over all time steps and compute the feature importance scores over the aggregated data.

This combination enables a precise analysis of the relevant features for the optimization of the LSTM model.

\section{Results}

\subsection{XAI results}

To evaluate the most important input features for our LSTM model, 10 training runs were performed with the parameters optimized by hyperparameter tuning. The average validation accuracy was 0.86, while the average F1 score was 0.89. The best run achieved a validation accuracy of 0.99 and an F1 score of 0.99, while the worst run achieved values of 0.68 for both metrics.

For the analysis of the most important features, the frequencies of the most frequently selected parameters from the three XAI methods SHAP, Grad-CAM, and LIME were counted. The 10 runs of each XAI method were used to rank the most frequently identified features. Based on these results, two reduced feature sets were defined: the 9 most frequent features and an optimized 6-feature set.

The `Total' column in Table \ref{tab:overall_xai}  represents the combined frequency across all three methods, summarizing all ten runs per method. The top six features were chosen as the target, since the original goal was to reduce the feature set to about one-third of the total, which was considered a reasonable starting point.

The table \ref{tab:overall_xai} shows the overall ranking of the 9 most important parameters across all methods, while the second table shows the detailed frequencies for SHAP, Grad-CAM and LIME.

As expected, SHAP, Grad-CAM, and LIME highlighted different key features. For example, Actual Clamping Force and End of Ejection Position were consistently important in all SHAP runs, while Injection Pressure was always among the top features in LIME.

\begin{table}[h]
\centering
\caption{Overall ranking of the 9 most important features across all XAI methods}
\begin{tabular}{|l|c|c|c|c|}
\hline
\textbf{Feature} & \textbf{Total} & \textbf{SHAP} & \textbf{Grad-CAM} & \textbf{LIME} \\ \hline
Injection Pressure  & 26        & 9 & 7  & 10 \\
Actual Clamping Force & 21      & 10 & 3 & 8 \\
Screw Position  & 21            & 6 & 8  & 7 \\
End of Ejection Position & 17   & 10 & 3 & 4 \\
Screw Torque  & 15              & 8 & 2  & 5 \\
Mold Position  & 12             & 8 & 2  & 2 \\
Screw Speed  & 11               & 3 & 7  & 5 \\
Contact Force  & 9              & 0 & 5  & 4 \\
Screw Velocity  & 8             & 0 & 5  & 3 \\
\hline
\end{tabular}
\label{tab:overall_xai}
\end{table}

The analysis of feature importance with SHAP, Grad-CAM and LIME shows clear differences in the weighting of the relevant input features. SHAP and LIME identified largely similar parameters, while Grad-CAM prioritized them differently. Particularly striking was the strong agreement between SHAP and LIME for the characteristics of Injection Pressure, Actual Clamping Force, and Screw Position, which were consistently classified as particularly relevant in both methods. Screw Torque and End of Ejection Position were also recognized as important factors for the prediction quality in both methods.

\begin{figure}[h]
\centering
\includegraphics[width=1\columnwidth]{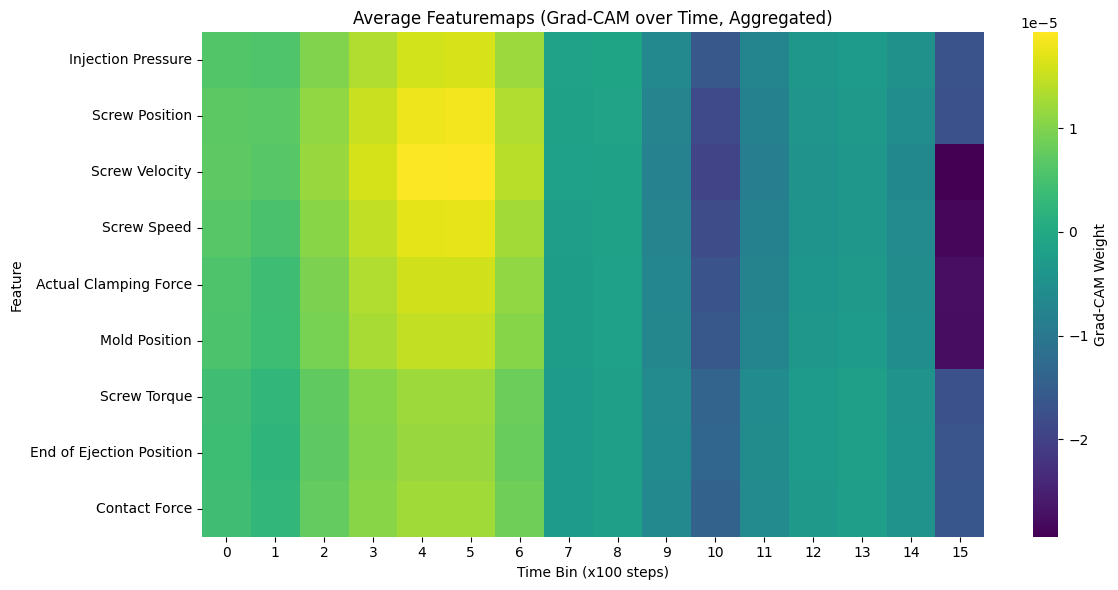}\
\vspace{1mm}
\includegraphics[width=1\columnwidth]{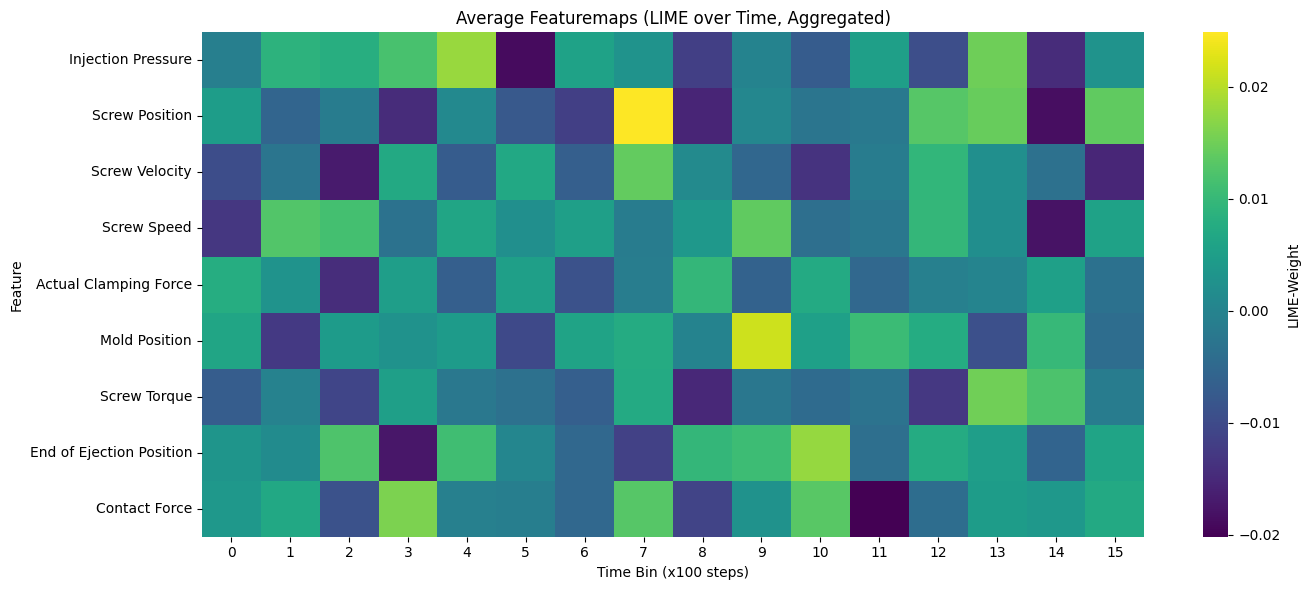}\
\vspace{1mm}
\includegraphics[width=1\columnwidth]{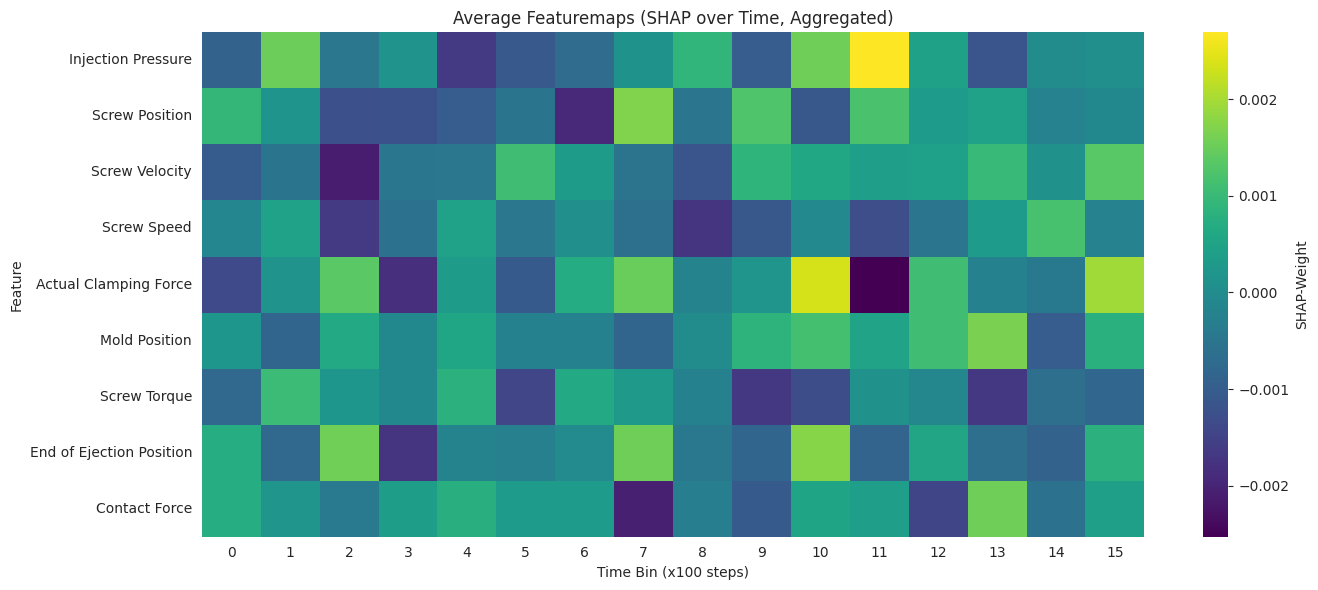}
\caption{Aggregated feature maps (Grad-CAM top, LIME middle, SHAP bottom) showing average feature relevance across time bins of the injection molding cycle.}
\label{fig:xai_heatmaps}
\end{figure}

By contrast, Grad-CAM placed greater emphasis on local movement variables, such as screw velocity, crosshead position, and cavity pressure, which received less attention in SHAP and LIME. In turn, SHAP and LIME consistently highlighted features like actual clamping force and injection pressure, which appeared less prominently in Grad-CAM rankings.
These differences suggest that SHAP and LIME integrate information across the full-time sequence, providing a broader view, while Grad-CAM focuses on localized activations within the LSTM model. To illustrate temporal patterns in feature relevance, Figure~\ref{fig:xai_heatmaps} shows average feature importance across the time intervals for SHAP, LIME, and Grad-CAM.
SHAP and LIME display significant temporal variation—for instance, SHAP highlights clamping force and mold position at specific time steps, while LIME emphasizes injection pressure and screw position later in the cycle. Grad-CAM, in contrast, yields a flatter temporal relevance, emphasizing motion parameters but lacking time-specific peaks, likely due to gradient averaging over time.

\subsection{Influence of feature reduction on model performance}

In order to investigate the influence of feature reduction on model performance, three variants of the LSTM model with 19, 9, and 6 features were tested over 10 runs. The results show that the reduction to 6 features led to a drop in performance, while the model with 9 features showed a better generalization capability.

The full model with 19 features achieved an average validation accuracy of 86\% and an F1 score of 89\%. After reducing the input to the 9 best features, the mean validation accuracy increased to 91\% and the mean F1 score to 92\%, suggesting that the feature selection improved the generalization of the model. Surprisingly, the 6-feature model was less stable, achieving a mean validation accuracy of 84\% and a mean F1 score of 86\%, despite the reduced input size. These results suggest that while feature reduction can improve performance, overly aggressive reduction can lead to a loss of relevant information.

The results in table \ref{tab:feature_reduction} show that the feature reduction does not necessarily lead to a drop in performance and that a targeted selection of relevant parameters can stabilize the model performance.

\begin{table}[h]
\centering
\caption{Mean validation accuracy and F1 score (mean ± std. dev.) over 10 runs, in percent.}
\begin{tabular}{|c|c|c|c|c|}
\hline
\textbf{Model} & \textbf{Val. Acc.} & \textbf{SD Acc} & \textbf{F1 Score} & \textbf{SD F1} \\ \hline
19 Features & 86.79 & 11.08 & 89.08 & 9.99 \\  
9 Features  & 91.33 & 5.79  & 92.52 & 5.34 \\  
6 Features  & 84.70 & 9.24  & 86.29 & 9.70 \\  
\hline
\end{tabular}
\label{tab:feature_reduction}
\end{table}


\subsection{Influence of feature reduction on inference time \& memory consumption}

In addition to model accuracy, the influence of feature reduction on inference time and memory consumption (VRAM usage) was also analyzed over all 10 models, which were generated during the XAI performance tests. For this purpose, measurements were performed on all used data, the training and the evaluation data set.

The results show that the model with 9 features has a 7\% faster inference time than the model with 19 features. The memory consumption was only minimally reduced by 0.4\%, since the VRAM consumption is mainly determined by the model architecture and less by the number of input features.

For the 6-feature model, the inference time and memory consumption were also measured, resulting in an inference time 13\% faster than with 19 parameters and a VRAM consumption also reduced by just 0.6\%. These values vary slightly due to different computational loads, but provide a solid basis for comparison. The inference results are shown in table \ref{tab:inference_memory}.

\begin{table}[h]
\centering
\caption{Impact of Feature Reduction on Inference Time and Memory Usage}
\begin{tabular}{|c|c|c|}
\hline
\textbf{Model} & \textbf{Mean inference Time (s)} & \textbf{Mean VRAM Usage (MB)} \\ \hline
19 Features  & 14.20 & 24,034.0 \\  
9 Features  & 13.26 & 23,928.1 \\  
6 Features  & 12.33 & 23,875.3 \\  
\hline
\end{tabular}
\label{tab:inference_memory}
\end{table}

\section{Conclusion}

We have shown that it is possible to identify the most influential parameters in the injection molding process using XAI. By applying the XAI methods SHAP, Grad-CAM and LIME to an LSTM model, the original 19 input factors were reduced to smaller sets of nine and six essential features. This reduction did not result in a significant loss of performance. Instead, the 9-feature model outperformed the full model, achieving an average validation accuracy of 91\% and an F1 score of 92\%, with a 5\% increase in validation accuracy and a 2\% increase in F1 score. The 6-feature model, while slightly less stable, still maintained a strong 84\% validation accuracy and 86\% F1 score. The 19-feature model had an average validation accuracy of 86\% and an F1 score of 89\%, demonstrating that feature selection can improve generalization while maintaining prediction quality.
Another goal of the study was to find out if parameter reduction can increase our inference efficiency. By reducing the number of input features, the 9-feature model showed a 7\% improvement in inference time, while the 6-feature model improved by 13\%. The VRAM savings were minimal because the memory consumption was largely determined by the model architecture rather than the number of input variables.

While the results are encouraging, the variability observed across training runs underscores the necessity of maintaining stable training conditions and employing diverse datasets to ensure the robustness of the system. However, this study successfully demonstrated that XAI can be used sucessfully to extract the most relevant features in an industrial injection molding process while maintaining or even improving model performance.
Another important finding was the potential to use the identified features to generate synthetic process data. Accurately modeling the many interdependencies required for simulating realistic data with 19 input parameters is highly complex. XAI-based feature reduction simplifies this task and mitigates the risk of unrealistic or miscalibrated correlations in synthetic data. This aligns with Lu et al. (Lu et al., 2023), who note that capturing correlations in high-dimensional input spaces is a major challenge. Our results suggest that classification performance can be maintained or even improved using only six to nine key features. This supports the idea that synthetic data based on fewer relevant parameters can effectively train models. Such reduction improves the feasibility and quality of synthetic data, thereby enhancing the robustness and practicality of AI-based quality prediction in industrial settings.
The findings of this study demonstrate the efficacy of applying XAI methods in industrial machine learning, thereby enhancing the interpretability and efficiency of an injection molding classification model. Furthermore, this study establishes a foundation for the more effective generation of synthetic data, which has the potential to substantially.

\bibliographystyle{IEEEtran}
\bibliography{references}


\end{document}